\pdfoutput=1

\def\year{2018}\relax
\documentclass[letterpaper]{article} 
\usepackage{aaai18}  
\usepackage{times}  
\usepackage{helvet}  
\usepackage{courier}  
\usepackage{url}  
\usepackage{graphicx}  
\usepackage{booktabs}
\usepackage{amsmath,amssymb,mathtools,stmaryrd,bm}
\usepackage{subcaption} 
\usepackage{algorithm,algorithmic}

\graphicspath{{./fig/}}

\newcommand{\rmd}{\mathrm{d}}
\newcommand{\Dataset}{\mathcal{D}}
\newcommand{\Cost}{\mathcal{L}}
\newcommand{\Obj}{\mathcal{G}}
\newcommand{\ngrad}{\tilde{\nabla}}

\newcommand{\wrt}{w.r.t.\ }

\begin{document}
%
\title{Dynamic Optimization of Neural Network Structures\\Using Probabilistic Modeling}
\author{
Shinichi Shirakawa\\
Yokohama National University\\
shirakawa-shinichi-bg@ynu.ac.jp\\
\And Yasushi Iwata\\
Yokohama National University\\
iwata-yasushi-ct@ynu.jp\\
\And Youhei Akimoto\\
Shinshu University\\
y\_akimoto@shinshu-u.ac.jp
}
\maketitle
\begin{abstract}
Deep neural networks (DNNs) are powerful machine learning models and have succeeded in various artificial intelligence tasks. Although various architectures and modules for the DNNs have been proposed, selecting and designing the appropriate network structure for a target problem is a challenging task. In this paper, we propose a method to simultaneously optimize the network structure and weight parameters during neural network training. We consider a probability distribution that generates network structures, and optimize the parameters of the distribution instead of directly optimizing the network structure. The proposed method can apply to the various network structure optimization problems under the same framework. We apply the proposed method to several structure optimization problems such as selection of layers, selection of unit types, and selection of connections using the MNIST, CIFAR-10, and CIFAR-100 datasets. The experimental results show that the proposed method can find the appropriate and competitive network structures.
\end{abstract}

\section{Introduction}
Deep neural networks (DNNs) have become a popular machine-learning model and seen great success in various tasks such as image recognition and natural language processing. To date, a variety of DNN models has been proposed. Considering the convolutional neural networks (CNNs) for visual object recognition as an example, a variety of deep and complex CNN models were developed, such as the VGG model \cite{Simonyan2015}, the residual networks (ResNets) \cite{He2016}, which have the skip connections, and the dense convolutional networks (DenseNets) \cite{Huang2017}. It is not easy for users to select an appropriate network structure including hyper-parameters, such as the depth of a network, the type of each unit, and the connection between layers, since the performance depends on tasks and data. However, the appropriate configuration of such structures is of importance for high performance of the DNNs. Therefore, developing efficient methods to optimize the structure of the DNNs is an important topic.

A popular approach to such a network structure optimization is to treat the network structure as the hyper-parameters of the DNNs and optimize them by a black-box optimization technique such as Bayesian optimization \cite{Snoek2012} or evolutionary algorithms \cite{Loshchilov2016}. Given a network configuration (hyper-parameter vector), the training is done for a certain period and the trained network is evaluated based on the accuracy or the loss for validation dataset. A black-box optimizer treats the hyper-parameter vector and the resulting accuracy/loss as the design variables and its objective/cost function. Recently, the methods for automatic network design that can construct more flexible network structures than the conventional hyper-parameter optimization approaches have been proposed. Zoph and Le defined the recurrent neural networks (RNNs) that generate neural network architectures for a target problem and found the state-of-the-art architectures by optimizing the RNN using the policy gradient method \cite{Zoph2017}. The works of \cite{Real2017,Suganuma2017} optimize the connections and types of layers by evolutionary algorithms to construct a high-performance CNN architecture. These methods succeeded in finding the state-of-the-art configurations of the DNNs. We view all these approaches as \textit{static optimization} of the network structures. The main disadvantage of static optimization is the efficiency, since it repeats the training with different network configurations until it finds a reasonable configuration.

A \textit{dynamic optimization} of the network structures, on the other side, learns the connection weights and the structure of a network simultaneously. A typical example is to represent the network structure parameters as the learnable parameters and optimize them by a stochastic gradient descent when carrying out the weight training \cite{Srinivas2016,Ba2013}. Srinivas and Babu \cite{Srinivas2016} introduce the Tri-State ReLU activation having differentiable parameters to prune the units and layers using back-propagation. Ba and Frey \cite{Ba2013} use a binary belief network overlaying a neural network to decide the dropout rate and jointly train two networks. The sparse and compact network structures can be dynamically learned by using regularization techniques \cite{Wen2016} as well. These methods require the loss function to be differentiable with respect to (w.r.t.) the structure parameters or they use heuristic optimization techniques. The dynamic optimization approach is computationally efficient since it optimizes the connection weights and the network structure within a single training loop, though it compromises the flexibility of the learnable structures compared with the static optimization approaches.


In this paper, we propose a general framework for dynamically optimizing the network structures and the connection weights simultaneously. To achieve more flexibility of learnable structures, we introduce a parametric distribution generating the network structure and treat the distribution parameters as the hyper-parameters. The objective function for the weights and hyper-parameters are defined by the expectation of the loss function under the distribution. Then, gradient based search algorithms can be applied. To demonstrate the flexibility and the efficiency of our framework, we consider the Bernoulli distributions in this paper and show that the proposed method can dynamically optimize various network structure parameters under the same framework. Our method is more computationally efficient than the static optimization approach and more flexible than the conventional dynamic optimization approach, such as directly optimizing structure parameters \cite{Srinivas2016,Ba2013}. We conduct four experiments: selection of layers, selection of activation functions, adaptation of stochastic network, and selection of connections. The experimental results show that the proposed method can find the appropriate and unusual network structures.

\section{Dynamic Network Structure Optimization}

\paragraph{Generic Framework}

In the following we consider the neural network $\phi(W, M)$ modeled by two parameter vectors, the vector $W \in \mathcal{W}$ consisting of the connection weights and the vector $M \in \mathcal{M}$ consisting of $d$ hyper-parameters that determine the structure of the network such as connectivity of each unit, type of activation function for each unit, and so on. The weights $W$ are in general real valued, and the structure parameters $M$ can live in an arbitrary space. Our original objective is to minimize the loss $\Cost(W, M)$, which is often defined as $\Cost (W, M) = \int_{\Dataset} l(z, W, M) p(z) \rmd z$, where $\Dataset$ and $l(z, W, M)$ indicate the dataset and the loss function for a given data $z$, respectively.

Let us consider a family of probability distributions $p_\theta(M)$ of $M$ parametrized by a real vector $\theta \in \Theta$. Instead of directly optimizing $\Cost(W, M)$, we consider to minimize the expected loss $\Obj(W, \theta)$ under $p_{\theta}(M)$, namely,
\begin{equation}
\Obj (W, \theta) = \int_{\mathcal{M}} \Cost (W, M) p_\theta(M) \rmd M \enspace,
\label{eq:objective}
\end{equation}
where $\rmd M$ is a reference measure on $\mathcal{M}$. Note that the minimizer $(W^*, \theta^*)$ of $\Obj$ admits the minimizer $(W^*, M^*)$ of $\Cost$ in the sense that $p_{\theta^*}$ will be concentrated at the minimizer $M^*$ of $\Cost$ as long as such a distribution is included in the given family of probability distributions. We remark that the domain $\mathcal{W}\times\Theta$ of $\Obj$ is continuous and the objective $\Obj$ is likely to be differentiable (one may be able to choose $p_\theta$ so that it will be), whereas $\Cost$ itself is not necessarily continuous since $\mathcal{M}$ can be discrete, as we consider in the following.

We optimize the parameters $W$ and $\theta$ based on by taking a gradient step and a natural gradient step, respectively, which are given by
\begin{align}
  \nabla_W \Obj (W, \theta) &= \int \nabla_W \Cost (W, M) p_\theta(M) \rmd M \label{eq:grad_w} ,
                              \\
\tilde{\nabla}_\theta \Obj (W, \theta) &= \int \Cost (W, M) \ngrad_{\theta} \ln p_\theta(M) p_\theta(M) \rmd M \label{eq:grad_theta} ,
\end{align}
where $\ngrad_{\theta} \ln p_\theta(M) = F^{-1}(\theta) \nabla_\theta \ln p_\theta(M)$ is the so-called natural gradient \cite{Amari1998} of the log-likelihood $\ln p_\theta$ and $F(\theta)$ is the Fisher information matrix of $p_\theta(M)$. Note that the natural gradient generally requires the estimation of the inverse Fisher information matrix in the standard machine learning setting (e.g., for $W$ update). However, we can analytically compute the natural gradient of the log-likelihood \wrt $\theta$ since we have full access to the distribution $p_\theta$. For example, any exponential family with sufficient statistics $T(M)$ with the expectation parameterization $\theta = E[T(M)]$ admits the natural gradient of the log-likelihood $\ngrad_{\theta}\ln p_\theta(M) = T(x) - \theta$. This setting is similar to those in the natural policy gradient method with parameter-based exploration (PGPE) \cite{Miyamae2010} for reinforcement learning and in the information geometric optimization algorithm (IGO) \cite{Ollivier2017} for simulation based black-box optimization. The natural gradient gives us a reasonable scaling of the gradient in our case, compared to the vanilla (Euclidean) gradient.

At each step of the training phase, we receive a mini-batch $\mathcal{Z}$ of $N$ training data $z_j$, i.e., $\mathcal{Z} = \{z_1, \dots, z_N\}$. The loss function $\Cost (W, M) = \int_{\Dataset} l(z, W, M) p(z) \rmd z$ is then approximated by the sample average of the loss $l(z_j, W, M)$. We shall write it as $\bar{\Cost}(W, M; \mathcal{Z})$. The cost in \eqref{eq:grad_theta} and the gradient of the cost in \eqref{eq:grad_w} are replaced with $\bar{\Cost}(W, M; \mathcal{Z})$ and $\nabla_{W}\bar{\Cost}(W, M; \mathcal{Z})$, respectively. In our situation, we need to estimate the cost and its gradient for each structure parameter $M_i$. We consider the following two different ways:
\begin{description}
\item[(a) same mini-batches] The same training data set $\mathcal{Z}$ is used for each $M_i$, namely,
\begin{equation}
\bar{\Cost} (W, M_i; \mathcal{Z}) = \frac{1}{N} \sum_{z \in \mathcal{Z}} l (z, W, M_i) \label{eq:approx_loss_same}
\end{equation}
\item[(b) different mini-batches] The training data set $\mathcal{Z}$ is decomposed into $\lambda$ subsets $\mathcal{Z}_i$ with equal number of data, $N / \lambda$, and each subset $\mathcal{Z}_i$ is used for each $M_i$, namely, 
\begin{equation}
  \bar{\Cost} (W, M_i; \mathcal{Z}_i) = \frac{\lambda}{N} \sum_{z \in \mathcal{Z}_i} l (z, W, M_i) \label{eq:approx_loss_diff}
\end{equation}
\end{description}
Letting $\bar{\Cost}(W, M_i)$ denote either \eqref{eq:approx_loss_same} or \eqref{eq:approx_loss_diff}, we obtain the Monte-Carlo approximation of the gradients as
\begin{align}
& \nabla_W \Obj (W, \theta) \approx \frac{1}{\lambda} \sum_{i=1}^{\lambda} \nabla_W \bar{\Cost} (W, M_i) \enspace, \label{eq:approx_grad_w} \\
& \ngrad_\theta \Obj (W, \theta) \approx \frac{1}{\lambda} \sum_{i=1}^{\lambda} \bar{\Cost} (W, M_i) \ngrad_\theta \ln p_\theta(M_i) \enspace. \label{eq:approx_grad_theta}
\end{align}
On one hand the latter \eqref{eq:approx_loss_diff} possibly has an advantage in computational time, since its mini-batch size for each network is $1 / \lambda$ times smaller. This advantage will disappear if GPU is capable of processing all the data of the original mini-batch in parallel. On the other hand, since the latter uses different batches to compute the loss of different networks, the resulting loss function may lead to a racing situation. From this optimization viewpoint, the former \eqref{eq:approx_loss_same} is preferred. These two variations are compared in experiment (I).

\paragraph{Instantiation with Bernoulli Distribution}

In the following we focus on the cases that the structure variables are binary, i.e., $m_k \in \{0, 1\}$ for each element $m_k$ of $M = (m_1, \dots, m_d)$. We consider the Bernoulli distribution $p_\theta$ as a law of $M$. The probability mass function is $p_{\theta}(M) = \prod \theta_k^{m_k}(1 - \theta_k)^{1 - m_k}$, where $\theta_k$ is the probability of each bit $m_k$ to be $1$. The parameter vector $\theta = (\theta_1, \dots, \theta_d)$ then lives in $\Theta = [0, 1]^{d}$. Since it is an exponential family with the expectation parameterization, the natural gradient of the log-likelihood $\ngrad \ln p_\theta (M)$ is given by $M - \theta$.

The parameters $W$ and $\theta$ are updated by taking the approximated gradient steps with learning rates (aka step-size) $\eta_W$ and $\eta_\theta$. Any stochastic gradient optimizer can be used for $W$ update as well as in the standard neural network training. However, since $\Theta$ is bounded, one needs to constrain $\theta$ so that it remains in $\Theta = [0, 1]^d$. To do so, a simple yet practically attractive treatment is to rescale the loss function at each training step. This is done by transforming the loss value $\bar{\Cost}(W, M_i)$ into the ranking based utility value $u_i$ \cite{Hansen2001,Yi2009} as
\begin{align}
\bar{\Cost}(W, M_i) \mapsto u_i = 
	\begin{cases}
		1 & (\mathrm{best} \; \lceil\lambda/4\rceil \; \mathrm{samples}) \\
		-1 & (\mathrm{worst} \; \lceil\lambda/4\rceil \; \mathrm{samples}) \\
		0 & (\mathrm{otherwise})
	\end{cases} .
\end{align}

With this utility transformation, the $\theta$ update reads
\begin{equation}
\theta^{t+1} = \theta^t + \frac{\eta_{\theta}}{\lambda} \sum_{i=1}^{\lambda} u_i (M_i - \theta^t) , 
\label{eq:update_rule}
\end{equation}
where $\eta_{\theta}$ is set to $1/(d \sum |u_i|)$ for all experiments. This way, we can guarantee for $\theta$ to stay in $\Theta$ with neither the adaptation of $\eta_{\theta}$ nor the constraint handling. Moreover, we restrict the range of $\theta$ within $[1/d, 1-1/d]$ to leave open the possibility of generating both values, i.e., we replace the value of $\theta$ by the boundary value if it will be updated beyond the boundary. The optimization procedure of the proposed method is displayed in Algorithm \ref{alg:procedure}.

\renewcommand{\algorithmicrequire}{\textbf{Input:}}
\renewcommand{\algorithmicensure}{\textbf{Output:}}
\begin{algorithm}[tbh]
\caption{Optimization procedure of the proposed method instantiated with Bernoulli distribution.}
\label{alg:procedure}
\begin{algorithmic}[1]
\REQUIRE Training data $\mathcal{D}$
\ENSURE Optimized parameters of $W$ and $\theta$
\renewcommand{\algorithmicrequire}{\textbf{Procedure:}}
\REQUIRE
\STATE Initialize the weights and Bernoulli parameters as $W^0$ and $\theta^0$
\STATE $t \leftarrow 0$
\WHILE{no stopping criterion is satisfied}
	\STATE Get $N$ mini-batch samples from $\mathcal{D}$
	\STATE Sample $M_0, \dots, M_\lambda$ from $p_{\theta^t}$
	\STATE Compute the loss using \eqref{eq:approx_loss_same} or \eqref{eq:approx_loss_diff}
	\STATE Update the weights to $W^{t+1}$ using \eqref{eq:approx_grad_w} by a SGD method
	\STATE Update the Bernoulli parameters to $\theta^{t+1}$ by \eqref{eq:update_rule}
	\STATE $t \leftarrow t + 1$
\ENDWHILE
\end{algorithmic}
\end{algorithm}

When we apply the trained network to test data, we have two options: the deterministic and stochastic predictions. The deterministic prediction indicates that we fix the random variables as $m_i = 0$ if $\theta_{i} < 0.5$ and $m_i = 1$ if $\theta_{i} \geq 0.5$, while the stochastic one averages the values of the model predictions using the samples from $p_\theta (M)$. The stochastic prediction requires high computational cost in proportion as the number of samples increases. We report the results of both predictions in the experiments and use $100$ samples for the stochastic prediction.

\paragraph{Relation to stochastic network models}
Since our method uses the stochastic network structures, we describe the relation to the stochastic networks such as Dropout \cite{Srivastava2014}, which stochastically zeros the output of hidden units during training to prevent overfitting. Also, other stochastic networks, DropConnect \cite{Wan2013} that drops the connections and the stochastic depth \cite{Huang2016a} that skips the layers in ResNets, were developed. Swapout \cite{Singh2016} is a generalization of Dropout and the stochastic depth, and it randomly chooses each unit behavior from four types: dropped, feedforward, skipped, or a residual network unit. These dropout techniques contribute to reducing the generalization error. The stochastic behavior is decided based on the Bernoulli distributions of typically $\theta = 0.5$. If the binary vector $M$ drawn from the Bernoulli distributions are used to decide whether each unit drops or not in our method, the method can be regarded as the adaptation of the dropout ratio. Therefore, our method can be also applied to the adaptation of the parameters of the existing stochastic network models.

\begin{table*}[tb]
\caption{Mean test errors ($\%$) over $30$ trials at the final iteration in the experiment of selection of layers. The values in parentheses denote the standard deviation.}
\label{tbl:err_layer_sel}
\centering
\begin{tabular}{llllll}
	\toprule
	 &  & \multicolumn{2}{c}{$\theta_{\mathrm{init}} = 0.5$} & \multicolumn{2}{c}{$\theta_{\mathrm{init}} = 0.969$} \\
	 &  & \multicolumn{1}{c}{Deterministic} & \multicolumn{1}{c}{Stochastic} & \multicolumn{1}{c}{Deterministic} & \multicolumn{1}{c}{Stochastic} \\
	\midrule
	AdaptiveLayer (a) & ($\lambda = 2$) & {\bf 2.200} ($0.125$)  & 2.218 ($0.135$) & {\bf 2.366} ($0.901$)  & $2.375$ ($0.889)$ \\
	AdaptiveLayer (b) & ($\lambda = 2$) & $15.69$ ($29.9$) & $15.67$ ($29.9$) & $35.26$ ($41.6$) & $35.27$ ($41.7$)\\
						& ($\lambda = 8$) & $2.406$ ($0.189$) & $2.423$ ($0.189$) & $65.74$ ($38.8$) & $65.74$ ($38.8$) \\
						& ($\lambda = 32$) & $2.439$ ($0.224$) & $2.453$ ($0.228$) & $80.59$ ($24.6$) & $80.60$ ($24.6$) \\
						& ($\lambda = 128$) & $2.394$ ($0.163$) & $2.405$ ($0.173$) & $80.58$ ($24.8$) & $80.58$ ($24.8$) \\
	\midrule
	StochasticLayer &  & \multicolumn{4}{c}{$4.704$ ($0.752$)} \\
\bottomrule
\end{tabular}
\end{table*}

\paragraph{Relation and difference to IGO}
Optimizing the parameters of the probability distribution in the proposed method is based on the IGO \cite{Ollivier2017}. We can view the IGO as a generalization of the specific estimation distribution algorithms (EDAs) \cite{Larranaga2001} such as the population based incremental learning (PBIL) \cite{Baluja1994} and the compact genetic algorithm (cGA) \cite{Harik1999} for discrete optimization. Moreover, it generalizes the covariance matrix adaptation evolution strategy (CMA-ES) \cite{Hansen2001,Hansen2003ec}, which is nowadays recognized as a stat-of-the-art black-box continuous optimizer. The update rule \eqref{eq:update_rule} is similar to the one in the cGA.

In the standard IGO and EDAs, the optimizer only updates the parameters of the distribution. On the contrary in this paper, the weight parameters of the neural network are simultaneously updated with a different mechanism (i.e., a stochastic gradient descent with momentum). Differently from applying IGO to update both parameters at the same time, we update the distribution parameters by IGO (i.e., natural gradient) and the weights are updated by using the gradient of the loss function, since the gradient is available and it leads to a faster learning compared to a direct search by IGO. 

From the viewpoint of updating the distribution parameters, i.e. optimizing the network structures, the landscape of the loss function dynamically changes at each algorithmic iteration because the weight parameters as well as a mini-batch change. This is the reason why we call the methods that optimize the both of structure and weight parameters at the same time \textit{dynamic structure optimization}.

\section{Experiments and Results}
\label{sec:experiment}
We apply our methodology to the following four situations: (I) selection of layers, (II) selection of activation functions, (III) adaptation of stochastic network, and (IV) selection of connections for densely connected CNNs. The algorithms are implemented by the Chainer framework \cite{Tokui2015} (version 1.23.0) on NVIDIA  Geforce GTX 1070 GPU for experiments (I) to (III) and on NVIDIA TITAN X GPU for experiment (IV). In all experiments, the SGD with a Nesterov momentum \cite{Sutskever2013} of $0.9$ and a weight decay of $10^{-4}$ is used to optimize the weight parameters. The learning rate is divided by $10$ at $1/2$ and $3/4$ of the maximum number of epochs. This setting is based on the literature \cite{He2016,Huang2017}.

\subsection{(I) Selection of Layers} \label{sec:exp_sel_layer}
\paragraph{Experimental setting}
The base network consists of $32$ fully connected hidden layers with $128$ units for each layer and the rectified linear unit (ReLU) \cite{Nair2010}. We use the MNIST handwritten digits dataset containing the 60,000 training examples and 10,000 test examples of $28 \times 28$ gray-scale images. The input and output layers correspond to the $784$ input pixel values and class labels ($0$ to $9$), respectively. We use the cross entropy error with softmax activation as the loss function $\Cost$. 

We use the binary vector $M=(m_1,\dots,m_d)$ to decide whether the processing of the corresponding layer is skipped: we skip the processing of $l$-th layer if $m_l = 0$. We re-connect the $(l+1)$-th layer with the  $(l-1)$-th layer when $m_l = 0$. More precisely, denoting the $l$-th layer's processing by $H_l$, the $(l+1)$-th layer's input vector becomes $X_{l+1} = H_l(X_l)$ if $m_l = 1$ and $X_{l+1} = X_l$ if $m_l = 0$. It is possible because the number of units in each layer is the same. The gradient $\nabla_{W}\bar{\Cost}(W, M_i)$ in \eqref{eq:approx_grad_w} is then computed in the straight-forward way, where the components of the gradient corresponding to the skipped layers are zero. Such skip processing is the same as the skip-forward defined in \cite{Singh2016}, and the number of $1$-bits in $M$ implies the number of hidden layers. To ensure the skip processing, we do not skip the first hidden layer and decide whether the second to 32-th hidden layers are skipped or not based on the binary vector. For this setting, the dimension of $M$ is $d = 31$.

The purpose of this experiment is to investigate the difference between the type of approximation of the loss, \eqref{eq:approx_loss_same} and \eqref{eq:approx_loss_diff}, and to check whether the proposed method can find the appropriate number of layers. With the neural network structure as mentioned above with fixed layer size and the following optimization setting, the training does not work properly when the number of layers is greater than $21$. Therefore, the proposed method needs to find less than 22 layers during the training. %
We denote the proposed methods using \eqref{eq:approx_loss_same} by AdaptiveLayer (a) and using \eqref{eq:approx_loss_diff} by AdaptiveLayer (b). We vary the parameter $\lambda$ as $\{2, 8, 32, 128\}$ for AdaptiveLayer (b) and use $\lambda = 2$ for AdaptiveLayer (a) and report the results using the deterministic and stochastic predictions mentioned above. The data sample size and the number of epochs are set to $N = 64$ and $100$ for AdaptiveLayer (a), respectively, and $N = 128$ and $200$ for other algorithms. The number of iterations is about $9 \times 10^4$ for all algorithms. At the beginning of training, we initialize the learning rate of SGD by $0.01$ and the Bernoulli parameters by $\theta_{\mathrm{init}} = 0.5$ or $\theta_{\mathrm{init}} = 1 - 1/31 \approx 0.968$ to verify the impact of $\theta$ initialization\footnote{The initialization with $\theta_{\mathrm{init}} = 1 - 1/31 \approx 0.968$ is an artificially poor initialization. We use this setting here only to check the impact of the initialization. We do not recommend tuning $\theta_{\mathrm{init}}$ at all, and it should be $\theta_{\mathrm{init}}=0.5$ which assumes no prior knowledge.}. We also run the method using fixed Bernoulli parameters of $0.5$ denoted by StochasticLayer to check the effectiveness of optimizing $\theta$. The experiments are conducted over $30$ trials with the same settings.

\begin{figure}[tb]
	\centering
	\includegraphics[width=0.85\linewidth]{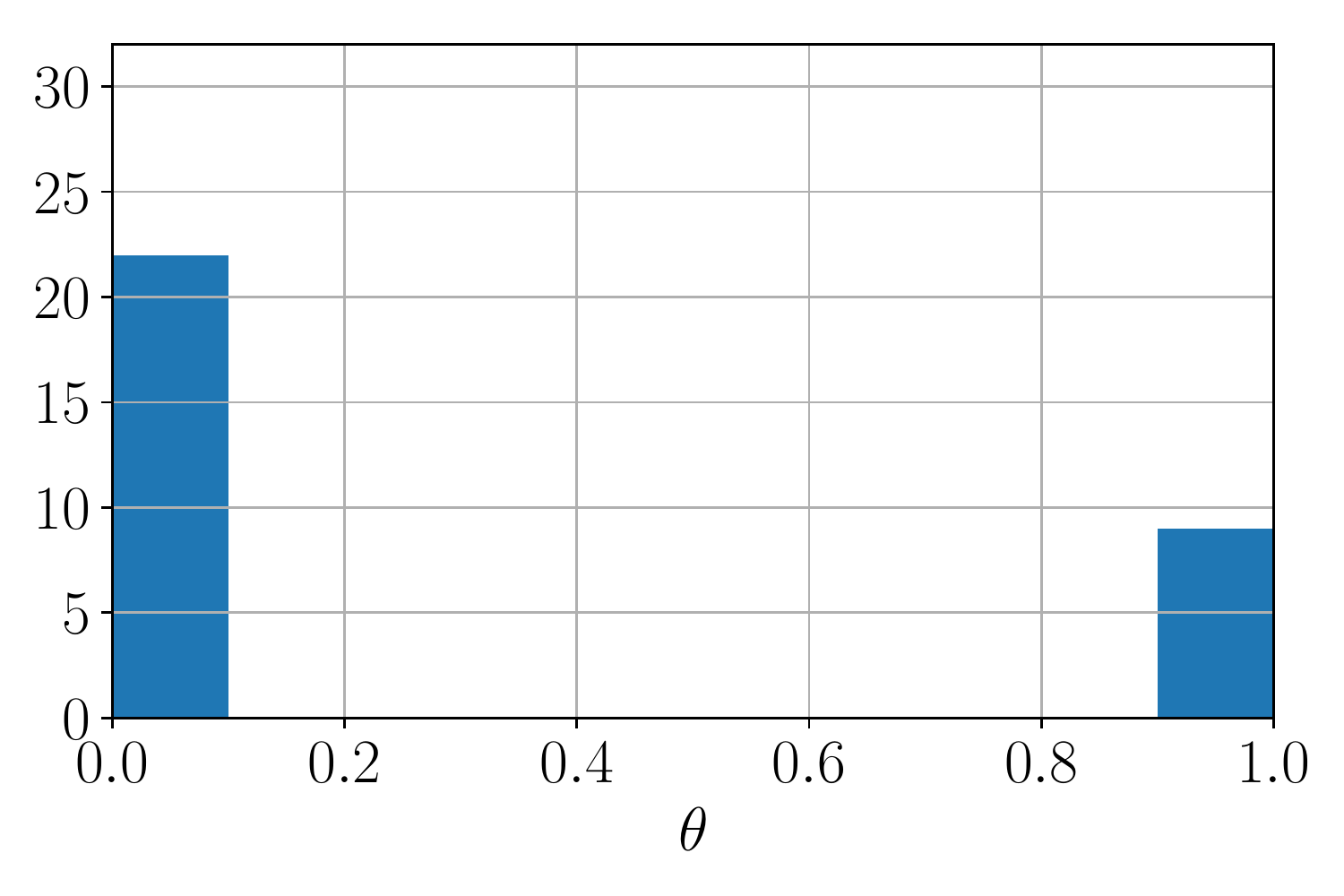}
	\caption{An example of histogram of $\theta$ obtained by AdaptiveLayer (a) at the final iteration.}
	\label{fig:hist_layer_sel}
\end{figure}

\begin{figure}[tb]
	\centering
	\includegraphics[width=0.85\linewidth]{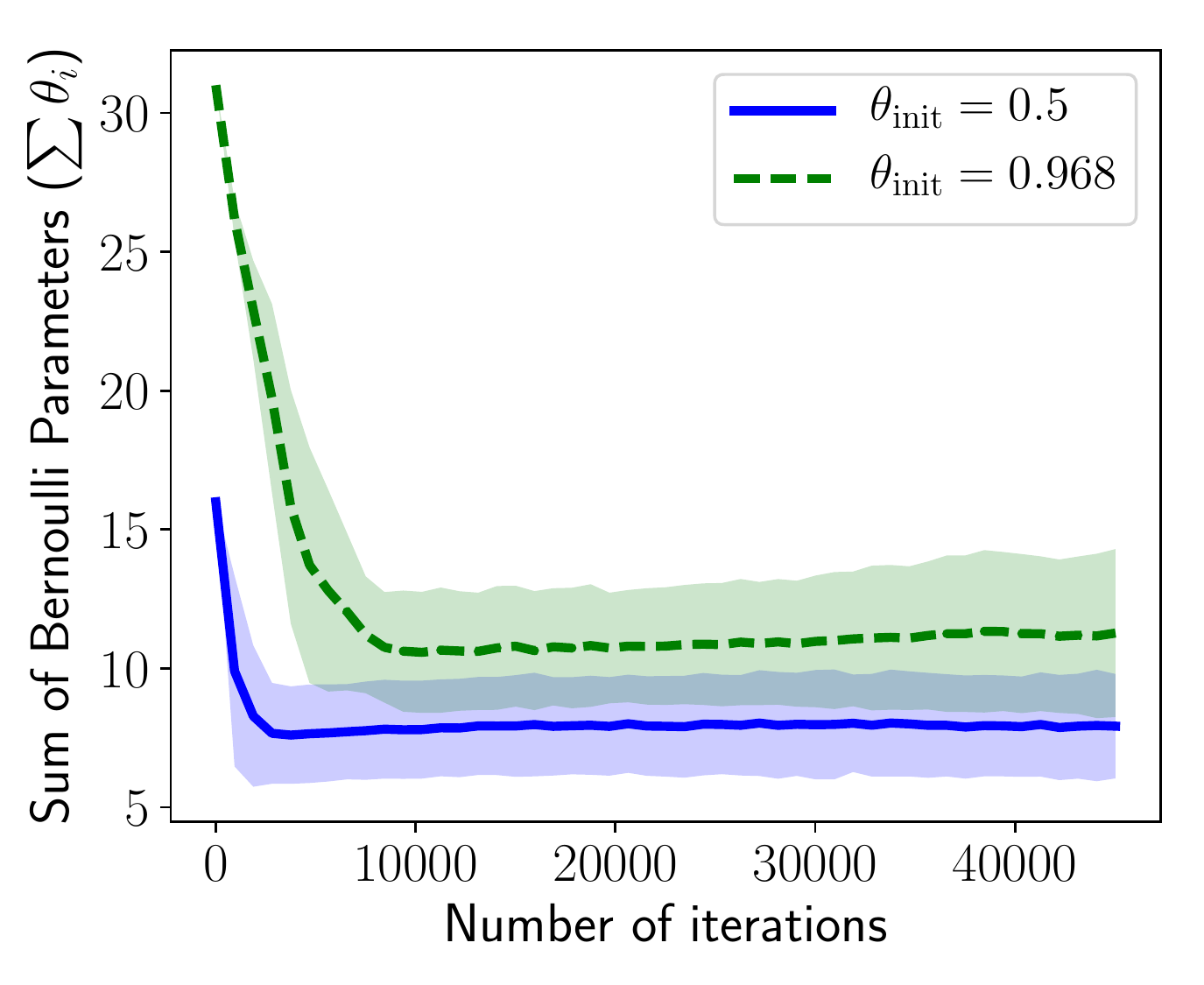}
	\caption{Transitions of the sum of the Bernoulli parameters ($\sum \theta_i$) using AdaptiveLayer (a) when we initialize by $\theta_\mathrm{init} = 0.5$ and $0.968$. The expected number of hidden layers is given by $\sum \theta_i + 1$. The first 45,000 iterations (about half of the maximum number of iterations) are plotted.}
	\label{fig:trans_theta_layer_sel}
\end{figure}

\paragraph{Result and discussion}
Table \ref{tbl:err_layer_sel} shows the test error of each method at the final iteration. We observe that AdaptiveLayer (a) shows the best performance among the proposed methods, and the performances of AdaptiveLayer (b) become significantly worse when the bad initialization ($\theta_{\mathrm{init}} \approx 0.968$) is used. One reason for this is that the loss approximation \eqref{eq:approx_loss_same} used in AdaptiveLayer (a) evaluates each sample of $M$ with the same mini-batch, and this leads to an accurate comparison between the samples of $M$. Comparing the deterministic and stochastic predictions, the performance differences are not significant because the values of $\theta$ distributed close to $0.0$ or $1.0$, as shown in Figure \ref{fig:hist_layer_sel}.

Figure \ref{fig:trans_theta_layer_sel} shows the transitions of the sum of the Bernoulli parameters ($\sum \theta_i$) for the first 45,000 iterations using AdaptiveLayer (a). The expected number of layers which is given by $\sum \theta_i + 1$ converges in between eight and ten. We observe that the values converge to the learnable number of layers at the early iteration, even in the case of the bad initial condition. In this experiments we observed no significant difference in computational time for AdaptiveLayer (a) and (b).

The test error of StochasticLayer is inferior to the most proposed methods; thus optimizing the Bernoulli parameters $\theta$ is effective. In our preliminary study, we found that the best number of layers was ten, whose test error is $1.778$ in our experimental setting. There was a run where the layer size converges to $10$, however, the final test error was inferior. From these observation, we conclude that the goodness of the proposed method is not to find the optimal network configuration, but to find a reasonable configuration within a single training loop. It will improve the convenience of the deep learning in practice. Based on the observation that the method using \eqref{eq:approx_loss_same} with $\lambda = 2$ showed the best performance, even in the case of the bad initial condition, we adopt this setting in the following experiments.

\begin{table*}[tb]
	\centering
	\caption{Mean test errors ($\%$) over $30$ trials at the final iteration in the experiment of selection of activation functions. The values in parentheses denote the standard deviation. The training time of a typical single run is reported.}
	\label{tbl:err_act_sel}
	\begin{tabular}{lccc}
		\toprule
		 &  Test error (Deterministic) & Test error (Stochastic) & Training time (min.)\\
		\midrule
		AdaptiveActivation  & $1.414$ ($0.054$) & {\bf 1.407} ($0.036$) & 255 \\
		StochasticActivation & \multicolumn{1}{c}{--} &  $1.452$ ($0.025$) & 204 \\
		ReLU     &  $1.609$ ($0.044$) & \multicolumn{1}{c}{--} & 120 \\
		tanh     &  $1.592$ ($0.069$) & \multicolumn{1}{c}{--} & 120 \\
		\bottomrule
	\end{tabular}
\end{table*}

\subsection{(II) Selection of Activation Functions} \label{sec:exp_sel_activation}
\paragraph{Experimental setting}
We use the binary vector $M$ to select the activation function for each unit. Different activation functions can be mixed in the same layer. The activation function of $i$-th unit is ReLU $F_\mathrm{relu}$ if $m_i = 1$ and the hyperbolic tangent $F_\mathrm{tanh}$ if $m_i = 0$. In other words, the activation function is defined as $m_i F_\mathrm{relu}(X_i) + (1 - m_i) F_\mathrm{tanh} (X_i)$, where $X_i$ denotes an input to the activation of the $i$-th unit. 

The base network structure used in this experiment consists of three fully connected hidden layers with 1,024 units for each layer. The number of activation functions to be decided is $d = 3072$. We use the MNIST dataset. In this experiment, we report the result of the method using \eqref{eq:approx_loss_same} with $\lambda = 2$ and denote it as AdaptiveActivation. We also run the method using the fixed Bernoulli parameters of $0.5$ and ones using the ReLU and hyperbolic tangent activations for all units; we denote them as StochasticActivation, ReLU, and tanh, respectively.

The data sample size and the number of epochs are set to $N = 64$ and 1,000 for AdaptiveActivation, respectively, and $N = 128$ and 2,000 for other algorithms. Note that the number of epochs is greater than the previous experiment. It is because the number of bits to be optimized (i.e., $3,072$) is significantly greater than the previous setting (i.e., $31$). We initialize the learning rate of SGD by $0.01$ and the Bernoulli parameters by $\theta_{\mathrm{init}} = 0.5$. The experiments are conducted over $10$ trials using the same settings.

\begin{figure}[tb]
	\centering
	\includegraphics[width=0.85\linewidth]{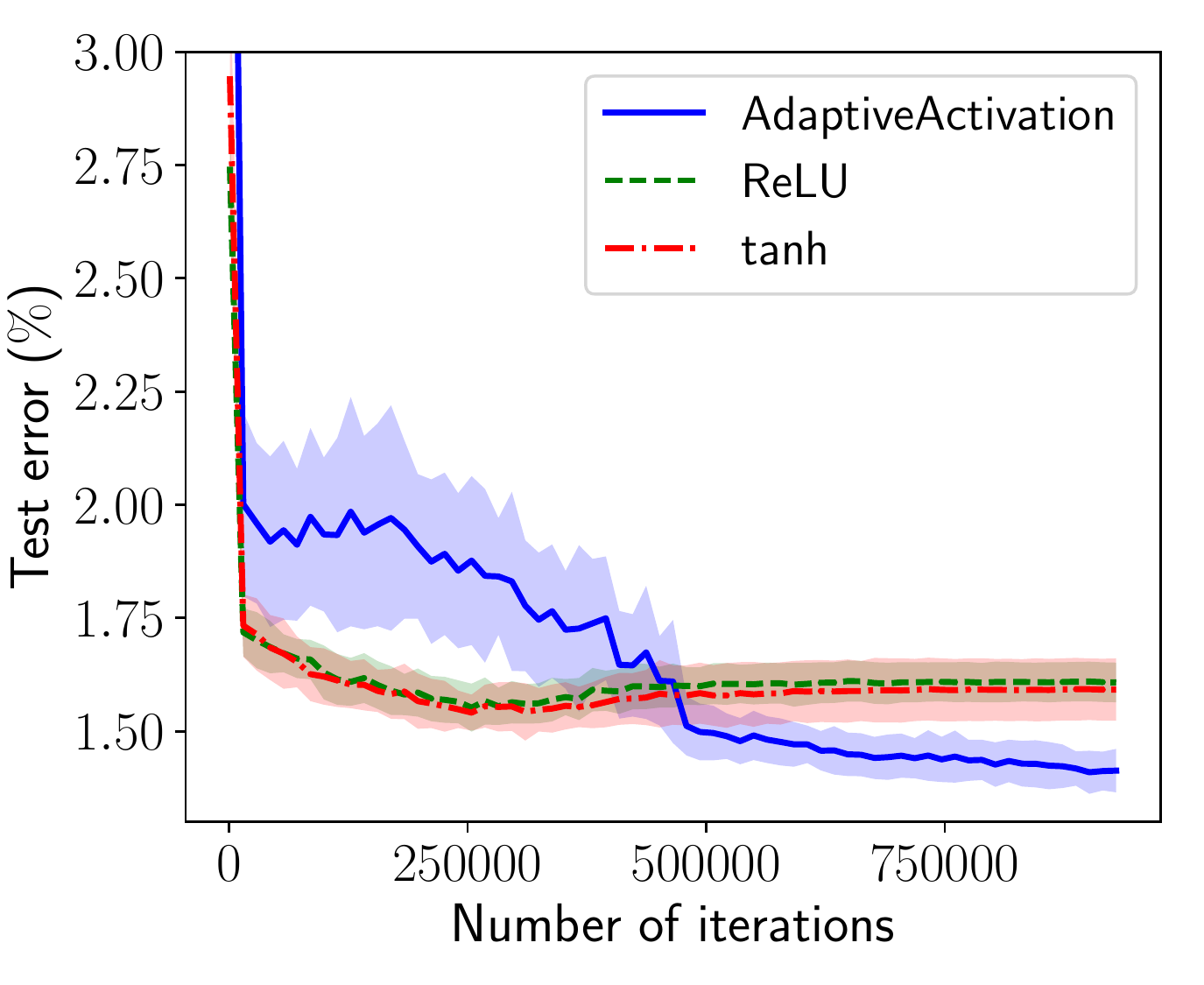}
	\caption{Transitions of the test errors ($\%$) of AdaptiveActivation (deterministic), ReLU, and tanh.}
	\label{fig:trans_err_act_sel}
\end{figure}

\begin{figure*}[tbh]
\centering
\begin{subfigure}[]{0.32\linewidth}
  \centering  
  \includegraphics[width=0.99\linewidth]{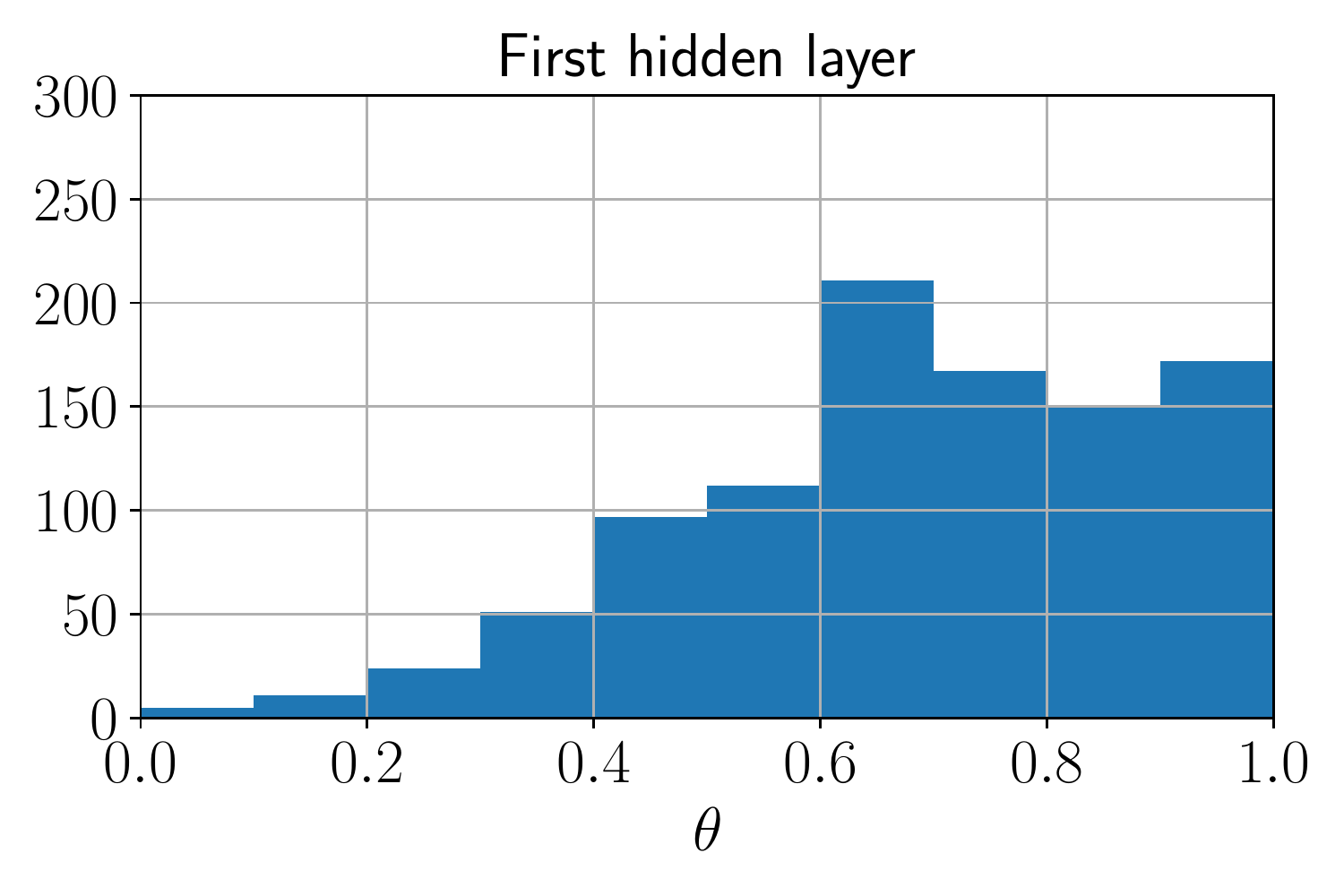}
\end{subfigure}
\begin{subfigure}[]{0.32\linewidth}
  \centering  
  \includegraphics[width=0.99\linewidth]{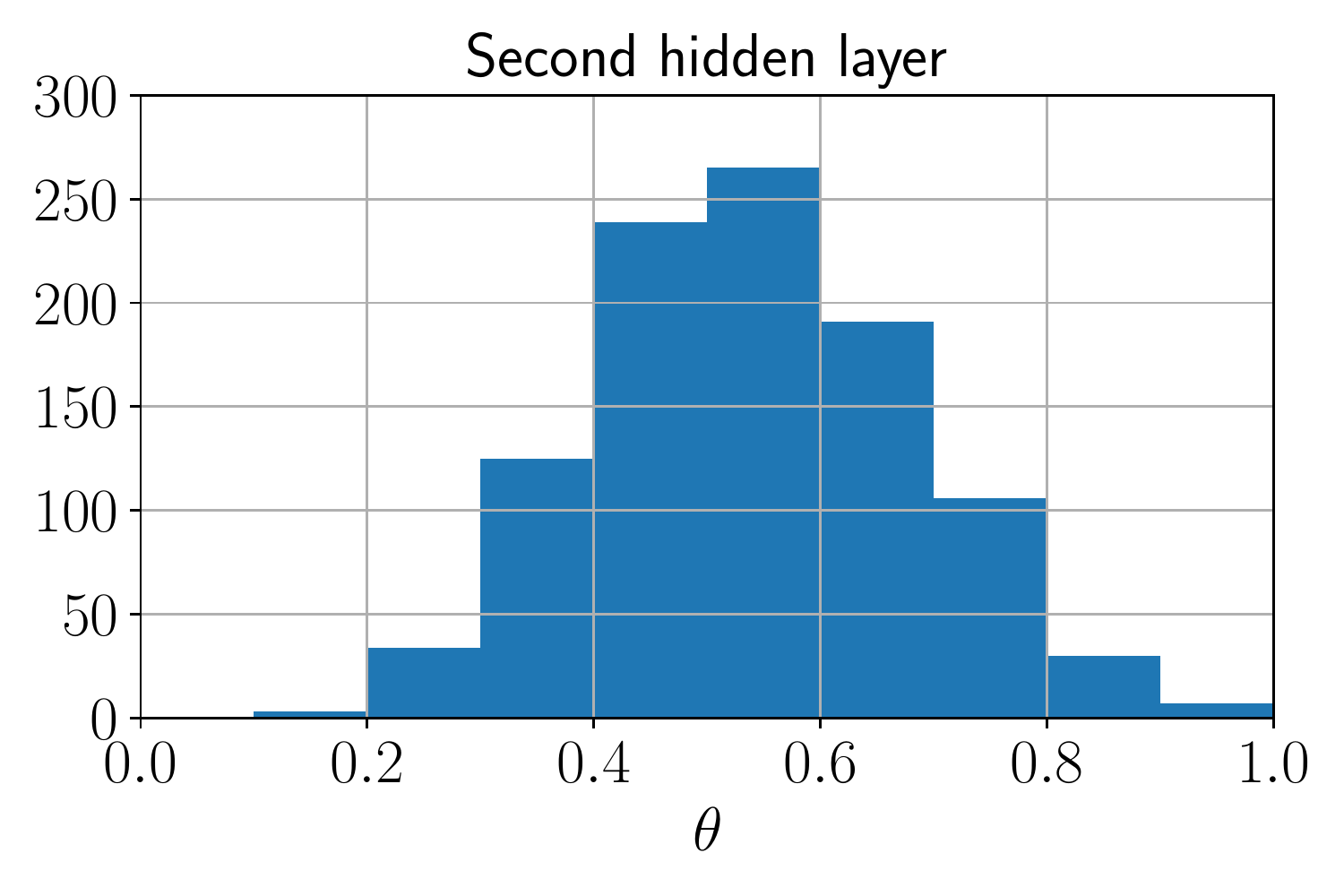}
\end{subfigure}
\begin{subfigure}[]{0.32\linewidth}
  \centering  
  \includegraphics[width=0.99\linewidth]{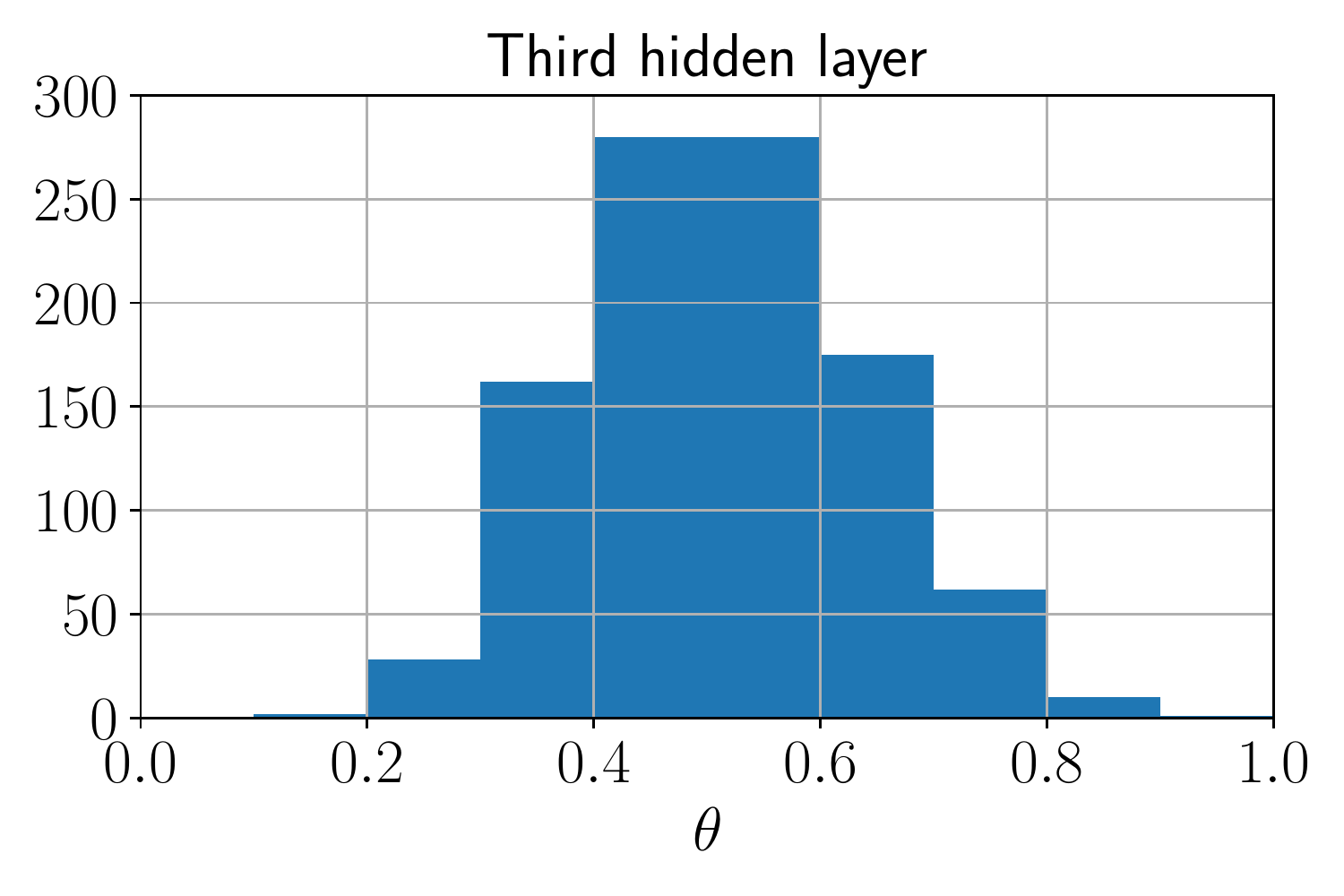}
\end{subfigure}
\caption{The histograms of $\theta$ in each layer obtained by AdaptiveActivation after training. The larger value of $\theta_i$ means that it tends to become ReLU. These histograms were created on a certain run, but the obtained histograms on the other runs are similar to this.}
\label{fig:hist_act_sel}
\end{figure*}

\paragraph{Result and discussion}
Table \ref{tbl:err_act_sel} shows the test error and training time of each algorithm. We observe that AdaptiveActivation (stochastic) outperforms StochasticActivation in which the Bernoulli parameters stay constant, suggesting that the optimization of such parameters by our method is effective. The predictive performance of AdaptiveActivation (deterministic) is competitive with StochasticActivation, but it is more computationally efficient than the stochastic prediction. In addition, the obtained networks by AdaptiveActivation have a better classification performance compared to both uniform activations: ReLU and hyperbolic tangent. Comparing the training time, we observe that the proposed method needs about twice the computational time for training compared to the fixed structured neural networks. Our method additionally requires the computation regarding the Bernoulli distributions (e.g., the equation \eqref{eq:update_rule}) and one to switch the structure. In our implementation, these are the reason of the increase in computational time. As our implementation is naive, the computational time may be reduced by a sophisticated implementation.

Figure \ref{fig:trans_err_act_sel} illustrates the transitions of the test errors of AdaptiveActivation (deterministic), ReLU, and tanh. We observe that the convergence of AdaptiveActivation is slow but achieves better results at the last iterations. More iterations are needed in our method to tune structure and weight parameters simultaneously.

Figure \ref{fig:hist_act_sel} shows an example of the histograms of $\theta$ in each layer after training. In our setting, the larger value of $\theta$ means that it tends to become ReLU. Interestingly, only the histogram of the first layer is biased toward ReLU. We have observed that the number of units with $\theta \geq 0.5$ increases to about 2,000 through training.

\subsection{(III) Adaptation of Stochastic Network} \label{sec:exp_ada_net}
\paragraph{Experimental setting}
Our proposed framework can be applied to optimize more than one types of hyper-parameters. To demonstrate this, we adapt the dropout ratio as well as the layer-skip ratio at the same time. We use the MNIST dataset in this experiment.

The network model is defined as follows. We consider a fully connected network consisting $L=10$ hidden layers with $U=1,024$ units for each layer as the base network. The configuration of the network is identified by $LU + (L - 1)$ binary parameters $M$. The first $L-1$ bits, denoted as $m^{\mathrm{layer}}_l$ for $l = 2,\dots,L$, determine whether $l$-th hidden layer is skipped (if $m^{\mathrm{layer}}_l = 0$) or not (if $m^{\mathrm{layer}}_l = 1$). The last $LU$ bits, denoted as $m^{\mathrm{unit}}_{li}$ for $l = 1,\dots,L$ and $i = 1,\dots,U$, determine whether the $i$-th unit of the $l$-th layer will be dropped (if $m^{\mathrm{unit}}_{li} = 0$) or not (if $m^{\mathrm{unit}}_{li} = 1$). The underlying probability distribution for $m^{\mathrm{layer}}_l$ is the Bernoulli distribution $p_{\theta^\mathrm{layer}_l}$, whereas the dropout mask $m^{\mathrm{unit}}_{li}$ for all $i = 1, \dots, U$ are drawn from the same Bernoulli distribution $p_{\theta^\mathrm{unit}_l}$. In other words, the dropout ratio is shared by units within a layer. Let the vector of the dropout mask for the $l$-th layer be denoted by $M^{\mathrm{unit}}_{l} = (m^{\mathrm{unit}}_{l1},\dots,m^{\mathrm{unit}}_{lU})$ and $p(M^{\mathrm{unit}}_{l}) = \prod_{i=1}^U p_{\theta^\mathrm{unit}_l}(m^{\mathrm{unit}}_{li})$. Then, the underlying distribution of $M$ is $p_\theta (M) = p(M^{\mathrm{unit}}_{1}) \prod_{l=2}^L p_{\theta^\mathrm{layer}_l}(m^{\mathrm{layer}}_l) p(M^{\mathrm{unit}}_{l})$ and the parameter vector is $\theta = (\theta^\mathrm{layer}_2, \dots, \theta^\mathrm{layer}_L, \theta^\mathrm{unit}_1, \dots, \theta^\mathrm{unit}_L) \in \mathbb{R}^{2L-1}$. 

Since the underlying probability distribution of $M$ is not anymore the independent Bernoulli model, the natural gradient of the log-likelihood is different from $M - \theta$. Yet, the natural gradient of the log-likelihood of our network model is easily derived as 
\begin{align}
& \ngrad_{\theta^\mathrm{layer}_l} \ln p_\theta(M) = m^{\mathrm{layer}}_{l} - \theta^\mathrm{layer}_{l} \enspace, \notag \\
& \ngrad_{\theta^\mathrm{unit}_l} \ln p_\theta(M) = \frac{1}{U}\sum_{i=1}^U m^{\mathrm{unit}}_{li} - \theta^\mathrm{unit}_{l} \enspace. \notag
\end{align}
(It demonstrates the generality of our methodology to some extent.) We use the same training parameters as used in the first experiment and report the result of the method using \eqref{eq:approx_loss_same} with $\lambda = 2$ and denote it as AdaptiveNet.

We employ a simple Bayesian optimization to the same problem to compare the computational cost with a static hyper-parameter optimization method. We use GPyOpt package (version 1.0.3, \url{http://github.com/SheffieldML/GPyOpt}) for the Bayesian optimization implementation and adopt the default parameter setting. The Bernoulli parameters of the stochastic network as mentioned above are optimized as the hyper-parameter. The problem dimension is $d=2L-1=19$ and the range of search space is $[1/d, 1 - 1/d]^d$. The training data is split into training and validation set in the ratio of nine to one; the validation set is used to evaluate a hyper-paremter after training the neural network with a candidate hyper-parameter. We fix the parameters of the Bernoulli distribution during the network training. After searching the hyper-parameter, we retrain the model using all training data and report the error for test data. For fair comparison, we include the vector of $(0.5, \dots, 0.5)$, which is the initial parameter of the proposed method, to the initial points for the Bayesian optimization. We use the same setting for the network training as used in the proposed method.

\begin{table}[tb]
	\centering
	\caption{Test errors ($\%$) and computational time of the proposed method (AdaptiveNet) and the Bayesian optimization (BO) with different budgets in the experiment of adaptation of stochastic network. The mean values over $30$ trials are reported in the proposed method, and the value in parentheses denotes the standard deviation. For the Bayesian optimization, the result of a single run is reported.}
	\label{tbl:err_stochastic_net}
	\begin{tabular}{lcc}
		\toprule
		 &  Test error (\%) & Time (hour)\\
		\midrule
		AdaptiveNet  & $1.645$ ($0.072$) & $1.01$ \\
		BO (budget=10) &  $1.780$ & $9.59$ \\
		BO (budget=20) &  $1.490$ & $18.29$ \\
		\bottomrule
	\end{tabular}
\end{table}

\begin{table*}[tb]
	\centering
	\caption{Test errors ($\%$) at the final iteration in the experiment of connection selection for DenseNets. The values in parentheses denote the standard deviation.}
	\label{tbl:err_connect_sel}
	\begin{tabular}{lcccc}
		\toprule
		& \multicolumn{2}{c}{CIFAR-10} & \multicolumn{2}{c}{CIFAR-100} \\
		& Deterministic & Stochastic & Deterministic & Stochastic \\
		\midrule
		AdaptiveConnection & $5.427$ ($0.167$) & $5.399$ ($0.153$) & $25.461$ ($0.408$) & {\bf 25.315} ($0.409$) \\
		Normal DenseNet ($40$ depth, $k=12$) & {\bf 5.050} ($0.147$) & \multicolumn{1}{c}{--} & $25.518$ ($0.380$) & \multicolumn{1}{c}{--} \\
		\bottomrule
	\end{tabular}
\end{table*}

\paragraph{Result and discussion}
Table \ref{tbl:err_stochastic_net} shows that the test errors of the stochastic networks obtained by the proposed method and the Bayesian optimization with different budgets, where budget indicates the number of hyper-parameters to be evaluated. We use the stochastic prediction with $100$ samples to calculate the test errors. Obviously, we observe that the computational time of the Bayesian optimization proportionally increases for the number of budgets while our method is more computationally efficient. The proposed method can find a competitive stochastic network with reasonable computational time. We observed that the networks obtained by the proposed method skip about five to seven layers and their units are not dropped with high probability. We also observed the same tendency for the network obtained by the Bayesian optimization. Although the Bayesian optimization could find a better configuration in this case within several ten budgets, it probably needs many budgets if the dimension of hyper-parameters increases such as in the setting of the experiment (II).

\subsection{(IV) Selection of Connections for DenseNets}
\paragraph{Experimental setting}
In this experiment, we use the dense convolutional networks (DenseNets) \cite{Huang2017}, a state-of-the-art architectures for image classification, as the base network structure. DenseNets contain several dense blocks and transition layers. The dense block comprises of $L_{\mathrm{block}}$ layers, each of which implements a non-linear transformation with a batch normalization (BN) \cite{Ioffe2015} followed by a ReLU activation and a $3 \times 3$ convolution. The size of the output feature-maps of each layer is the same as that of the input feature-maps. Let $k$ be the number of output feature-maps of each layer, called growth rate; the $l$-th layer in the dense block receives $k (l - 1) + k_0$ feature-maps, where $k_0$ indicates the number of input feature-maps to the dense block. Thus, the number of output feature-maps of the dense block is $k L_{\mathrm{block}} + k_0$. The transition layer is located between the dense blocks and consists of a batch normalization, a ReLU activation, and an $1 \times 1$ convolutional layer followed by a $2 \times 2$ average pooling layer. The detailed architecture of DenseNets can be found in \cite{Huang2017}.

We decide the existence of the connections between layers in each dense block according to the binary vector $M$. Namely, we remove the connection when the corresponding bit equals zero. Let us denote the $k$-th layer's output feature-maps by $Y_{k}$; then, the input feature-maps to the $l$-th layer is computed by $(m_{p} Y_0, \dots, m_{p + l - 1} Y_{l-1})$, where $p = l(l-1)/2$. We use the most simple DenseNet consisting 40 depth ($k = 12$ and $L_{\mathrm{block}} = 12$) reported in \cite{Huang2017} as the base network structure, containing three dense blocks and two transition layers. For this setting, the dimension of $M$ becomes $d = 273$.

In this experiment, we use the CIFAR-10 and CIFAR-100 datasets in which the numbers of classes are $10$ and $100$, respectively. The numbers of training and test images are 50,000 and 10,000, respectively, and the size of the images is $32 \times 32$. We normalize the data using the per-channel means and the standard deviations in the preprocessing. We use the data augmentation method based on \cite{He2016,Huang2017}: padding 4 pixels on each side followed by choosing a random $32 \times 32$ crop from the padded image and random horizontal flips on the cropped $32 \times 32$ image.
We report the results of the method using \eqref{eq:approx_loss_same} with $\lambda = 2$ (AdaptiveConnection) and also run the normal DenseNet for comparison. The data sample size and the number of epochs are set to $N = 32$ and 300 for AdaptiveConnection, respectively, and $N = 64$ and 600 epochs for the normal DenseNet. We initialize the weight parameters using the method described in \cite{He2015}, and the learning rate of SGD and the initial Bernoulli parameters by $0.1$ and $0.5$, respectively. We conduct the experiments with same settings over $20$ and $5$ trials for AdaptiveConnection and normal DenseNet, respectively. 

\paragraph{Result and discussion}
Table \ref{tbl:err_connect_sel} shows the test errors of AdaptiveConnection and the normal DenseNet at the final iteration. In this case, the stochastic prediction is slightly better than the deterministic one, but the difference is not significant. The difference of the predictive performances between AdaptiveConnection and the normal DenseNet is not significant for the CIFAR-100 datasets, whereas AdaptiveConnection is inferior for the CIFAR-10 dataset. We, however, observed that the obtained Bernoulli parameters are distributed to be close to $0.0$ or $1.0$, as in Figure \ref{fig:hist_layer_sel}. We observed that about 70 connections are removed with high probability for both datasets. Counting the weight parameters of those removed connections, we found that it can reduce about $10\%$ of the weight parameters without suffering from performance deterioration for CIFAR-100.

\section{Conclusion}
In this paper, we proposed a methodology that dynamically and indirectly optimizes the network structure parameters by using probabilistic models. We instantiated the proposed method using the Bernoulli distributions and simultaneously optimized their parameters and network weights. We conducted experiments where we optimized four different network components: the layer skips, activation functions, layer skips and unit dropouts, and connections. We observed that the proposed method could find the learnable layer size and the appropriate mix rate of the activation functions. We also showed that our method can dynamically optimize more than one type of hyper-parameters and obtain the competitive results with a reasonable training time. In the experiment of connection selection for DenseNets, the proposed method have shown the competitive results with a smaller number of connections.

The proposed method is computationally more efficient than static structure optimization in general, which is validated in the experiment (III) (Table 3). The static optimization method such as a Bayesian optimization may find better hyper-parameter configuration, but it takes a lot more time. This is also observed in Table 3.

  The existing dynamic structure optimization methods need to parameterize the network structure by differentiable parameters to optimize them within a standard stochastic gradient descent framework, whereas the proposed method can optimize the network structure that are not necessarily differentiable through parametric probability distributions. Although this paper focuses on the Bernoulli distributions (0/1 bits), the proposed framework can be used with other distributions such as categorical distributions which represent several categorical variables (A/B/C/...). Indeed, it is rather easy to derive the update of the distribution parameters if the distributions are in exponential families. Since it is difficult to design a model to represent categorical variables by differentiable parameters, the proposed framework is more flexible than the existing dynamic optimization methods in the sense that it can handily treat a wider range of structural optimization problems. 

One direction of future work is to extend the proposed method to treat variables other than binary variables, i.e., categorical variables, and to optimize a larger and more complex networks. Another direction of future work is to introduce a prior distribution for $\theta$; one can incorporate the regularization term to obtain sparse and compact representation through the prior distribution of $\theta$.

\section{Acknowledgments}
This work is partially supported by the SECOM Science and Technology Foundation.

\bibliography{aaai2018-shirakawa}

\begin{thebibliography}{}

\bibitem[\protect\citeauthoryear{Amari}{1998}]{Amari1998}
Amari, S.
\newblock 1998.
\newblock Natural gradient works efficiently in learning.
\newblock {\em Neural Computation} 10(2):251--276.

\bibitem[\protect\citeauthoryear{Ba and Frey}{2013}]{Ba2013}
Ba, J., and Frey, B.
\newblock 2013.
\newblock Adaptive dropout for training deep neural networks.
\newblock In {\em Advances in Neural Information Processing Systems 26 (NIPS
  2013)},  3084--3092.

\bibitem[\protect\citeauthoryear{Baluja}{1994}]{Baluja1994}
Baluja, S.
\newblock 1994.
\newblock Population-based incremental learning: A method for integrating
  genetic search based function optimization and competitive learning.
\newblock Technical Report Tech Rep CMU-CS-94-163, Carnegie Mellon University.

\bibitem[\protect\citeauthoryear{Hansen and Ostermeier}{2001}]{Hansen2001}
Hansen, N., and Ostermeier, A.
\newblock 2001.
\newblock Completely derandomized self-adaptation in evolution strategies.
\newblock {\em Evolutionary Computation} 9(2):159--195.

\bibitem[\protect\citeauthoryear{Hansen, M\"{u}ller, and
  Koumoutsakos}{2003}]{Hansen2003ec}
Hansen, N.; M\"{u}ller, S.~D.; and Koumoutsakos, P.
\newblock 2003.
\newblock Reducing the time complexity of the derandomized evolution strategy
  with covariance matrix adaptation ({CMA-ES}).
\newblock {\em Evolutionary Computation} 11(1):1--18.

\bibitem[\protect\citeauthoryear{Harik, Lobo, and Goldberg}{1999}]{Harik1999}
Harik, G.~R.; Lobo, F.~G.; and Goldberg, D.~E.
\newblock 1999.
\newblock The compact genetic algorithm.
\newblock {\em IEEE Transactions on Evolutionary Computation} 3(4):287--297.

\bibitem[\protect\citeauthoryear{He \bgroup et al\mbox.\egroup }{2015}]{He2015}
He, K.; Zhang, X.; Ren, S.; and Sun, J.
\newblock 2015.
\newblock Delving deep into rectifiers: Surpassing human-level performance on
  imagenet classification.
\newblock In {\em Proceedings of the 2015 IEEE International Conference on
  Computer Vision (ICCV 2015)},  1026--1034.

\bibitem[\protect\citeauthoryear{He \bgroup et al\mbox.\egroup }{2016}]{He2016}
He, K.; Zhang, X.; Ren, S.; and Sun, J.
\newblock 2016.
\newblock Deep residual learning for image recognition.
\newblock In {\em Proceedings of the 2016 IEEE Conference on Computer Vision
  and Pattern Recognition (CVPR 2016)},  770--778.

\bibitem[\protect\citeauthoryear{Huang \bgroup et al\mbox.\egroup
  }{2016}]{Huang2016a}
Huang, G.; Sun, Y.; Liu, Z.; Sedra, D.; and Weinberger, K.~Q.
\newblock 2016.
\newblock Deep networks with stochastic depth.
\newblock In {\em Proceedings of the 14th European Conference on Computer
  Vision (ECCV 2016)}, volume 9908 of {\em LNCS},  646--661.
\newblock Springer.

\bibitem[\protect\citeauthoryear{Huang \bgroup et al\mbox.\egroup
  }{2017}]{Huang2017}
Huang, G.; Liu, Z.; van~der Maaten, L.; and Weinberger, K.~Q.
\newblock 2017.
\newblock Densely connected convolutional networks.
\newblock In {\em Proceedings of the 2017 IEEE Conference on Computer Vision
  and Pattern Recognition (CVPR 2017)},  4700--4708.

\bibitem[\protect\citeauthoryear{Ioffe and Szegedy}{2015}]{Ioffe2015}
Ioffe, S., and Szegedy, C.
\newblock 2015.
\newblock Batch normalization: Accelerating deep network training by reducing
  internal covariate shift.
\newblock In {\em Proceedings of the 32nd International Conference on Machine
  Learning (ICML 2015)}, volume~37,  448--456.
\newblock PMLR.

\bibitem[\protect\citeauthoryear{Larra\~{n}aga and
  Lozano}{2001}]{Larranaga2001}
Larra\~{n}aga, P., and Lozano, J.~A.
\newblock 2001.
\newblock {\em Estimation of Distribution Algorithms: A New Tool for
  Evolutionary Computation}.
\newblock Kluwer Academic Publishers.

\bibitem[\protect\citeauthoryear{Loshchilov and Hutter}{2016}]{Loshchilov2016}
Loshchilov, I., and Hutter, F.
\newblock 2016.
\newblock {CMA-ES} for hyperparameter optimization of deep neural networks.
\newblock {\em arXiv preprint}.

\bibitem[\protect\citeauthoryear{Miyamae \bgroup et al\mbox.\egroup
  }{2010}]{Miyamae2010}
Miyamae, A.; Nagata, Y.; Ono, I.; and Kobayashi, S.
\newblock 2010.
\newblock Natural policy gradient methods with parameter-based exploration for
  control tasks.
\newblock In {\em Advances in Neural Information Processing Systems 23 (NIPS
  2010)},  1660--1668.

\bibitem[\protect\citeauthoryear{Nair and Hinton}{2010}]{Nair2010}
Nair, V., and Hinton, G.~E.
\newblock 2010.
\newblock Rectified linear units improve restricted boltzmann machines.
\newblock In {\em Proceedings of the 27th International Conference on Machine
  Learning (ICML 2010)},  807--814.

\bibitem[\protect\citeauthoryear{Ollivier \bgroup et al\mbox.\egroup
  }{2017}]{Ollivier2017}
Ollivier, Y.; Arnold, L.; Auger, A.; and Hansen, N.
\newblock 2017.
\newblock Information-geometric optimization algorithms: A unifying picture via
  invariance principles.
\newblock {\em Journal of Machine Learning Research} 18:1--65.

\bibitem[\protect\citeauthoryear{Real \bgroup et al\mbox.\egroup
  }{2017}]{Real2017}
Real, E.; Moore, S.; Selle, A.; Saxena, S.; Suematsu, Y.~L.; Tan, J.; Le,
  Q.~V.; and Kurakin, A.
\newblock 2017.
\newblock Large-scale evolution of image classifiers.
\newblock In {\em Proceedings of the 34th International Conference on Machine
  Learning (ICML 2017)}, volume~70,  2902--2911.
\newblock PMLR.

\bibitem[\protect\citeauthoryear{Simonyan and Zisserman}{2015}]{Simonyan2015}
Simonyan, K., and Zisserman, A.
\newblock 2015.
\newblock Very deep convolutional networks for large-scale image recognition.
\newblock In {\em Proceedings of the 3rd International Conference on Learning
  Representations (ICLR 2015)}.

\bibitem[\protect\citeauthoryear{Singh, Hoiem, and Forsyth}{2016}]{Singh2016}
Singh, S.; Hoiem, D.; and Forsyth, D.
\newblock 2016.
\newblock Swapout: Learning an ensemble of deep architectures.
\newblock In {\em Advances in Neural Information Processing Systems 29 (NIPS
  2016)},  28--36.

\bibitem[\protect\citeauthoryear{Snoek, Larochelle, and
  Adams}{2012}]{Snoek2012}
Snoek, J.; Larochelle, H.; and Adams, R.~P.
\newblock 2012.
\newblock Practical {Bayesian} optimization of machine learning algorithms.
\newblock In {\em Advances in Neural Information Processing Systems 25 (NIPS
  2012)},  2951--2959.

\bibitem[\protect\citeauthoryear{Srinivas and Babu}{2016}]{Srinivas2016}
Srinivas, S., and Babu, R.~V.
\newblock 2016.
\newblock Learning neural network architectures using backpropagation.
\newblock In {\em Proceedings of the British Machine Vision Conference (BMVC
  2016)}.

\bibitem[\protect\citeauthoryear{Srivastava \bgroup et al\mbox.\egroup
  }{2014}]{Srivastava2014}
Srivastava, N.; Hinton, G.; Krizhevsky, A.; Sutskever, I.; and Salakhutdinov,
  R.
\newblock 2014.
\newblock Dropout: A simple way to prevent neural networks from overfitting.
\newblock {\em Journal of Machine Learning Research} 15:1929--1958.

\bibitem[\protect\citeauthoryear{Suganuma, Shirakawa, and
  Nagao}{2017}]{Suganuma2017}
Suganuma, M.; Shirakawa, S.; and Nagao, T.
\newblock 2017.
\newblock A genetic programming approach to designing convolutional neural
  network architectures.
\newblock In {\em Proceedings of the Genetic and Evolutionary Computation
  Conference 2017 (GECCO 2017)},  497--504.

\bibitem[\protect\citeauthoryear{Sutskever \bgroup et al\mbox.\egroup
  }{2013}]{Sutskever2013}
Sutskever, I.; Martens, J.; Dahl, G.; and Hinton, G.
\newblock 2013.
\newblock On the importance of initialization and momentum in deep learning.
\newblock In {\em Proceedings of the 30th International Conference on Machine
  Learning (ICML 2013)}, volume~28,  1139--1147.
\newblock PMLR.

\bibitem[\protect\citeauthoryear{Tokui \bgroup et al\mbox.\egroup
  }{2015}]{Tokui2015}
Tokui, S.; Oono, K.; Hido, S.; and Clayton, J.
\newblock 2015.
\newblock Chainer: a next-generation open source framework for deep learning.
\newblock In {\em Proceedings of the Workshop on Machine Learning Systems
  (LearningSys) in the 29th Annual Conference on Neural Information Processing
  Systems (NIPS 2015)},  1--6.

\bibitem[\protect\citeauthoryear{Wan \bgroup et al\mbox.\egroup
  }{2013}]{Wan2013}
Wan, L.; Zeiler, M.; Zhang, S.; Cun, Y.~L.; and Fergus, R.
\newblock 2013.
\newblock Regularization of neural networks using {DropConnect}.
\newblock In {\em Proceedings of the 30th International Conference on Machine
  Learning (ICML 2013)}, volume~28,  1058--1066.
\newblock PMLR.

\bibitem[\protect\citeauthoryear{Wen \bgroup et al\mbox.\egroup
  }{2016}]{Wen2016}
Wen, W.; Wu, C.; Wang, Y.; Chen, Y.; and Li, H.
\newblock 2016.
\newblock Learning structured sparsity in deep neural networks.
\newblock In {\em Advances in Neural Information Processing Systems 29 (NIPS
  2016)},  2074--2082.

\bibitem[\protect\citeauthoryear{Yi \bgroup et al\mbox.\egroup }{2009}]{Yi2009}
Yi, S.; Wierstra, D.; Schaul, T.; and Schmidhuber, J.
\newblock 2009.
\newblock Stochastic search using the natural gradient.
\newblock In {\em Proceedings of the 26th International Conference on Machine
  Learning (ICML 2009)},  1161--1168.

\bibitem[\protect\citeauthoryear{Zoph and Le}{2017}]{Zoph2017}
Zoph, B., and Le, Q.~V.
\newblock 2017.
\newblock Neural architecture search with reinforcement learning.
\newblock In {\em Proceedings of the 5th International Conference on Learning
  Representations (ICLR 2017)}.

\end{thebibliography}
\bibliographystyle{aaai}

\end{document}